\documentclass{article}

\usepackage[utf8]{inputenc} \usepackage[T1]{fontenc}    \usepackage{hyperref}       \usepackage{url}            \usepackage{booktabs}       \usepackage{amsfonts}       \usepackage{nicefrac}       \usepackage{color}
\usepackage{epsfig}
\usepackage{amsthm}
\usepackage{amsmath}
\usepackage{float}
\usepackage{graphicx}
\usepackage{wrapfig}
\usepackage{makecell}
\usepackage{caption}
\usepackage{subcaption}
\usepackage{bm}
\usepackage{multirow}
\usepackage{microtype}
\usepackage{mathtools}

\newcommand{\nop}[1]{}

\DeclareMathOperator*{\argmax}{arg\,max}
\DeclareMathOperator*{\argmin}{arg\,min}

\newtheorem{proposition}{Proposition}

\usepackage[accepted]{icml2020}

\begin{document}

\twocolumn[
\icmltitle{Differentiable Product Quantization for End-to-End Embedding Compression}

\icmlsetsymbol{equal}{*}

\begin{icmlauthorlist}
\icmlauthor{Ting Chen}{goo}
\icmlauthor{Lala Li}{goo}
\icmlauthor{Yizhou Sun}{ucla}
\end{icmlauthorlist}

\icmlaffiliation{goo}{Google Research}
\icmlaffiliation{ucla}{University of California, Los Angeles}

\icmlcorrespondingauthor{Ting Chen}{iamtingchen@google.com}

\icmlkeywords{Machine Learning, ICML}

\vskip 0.3in
]

\printAffiliationsAndNotice{}

\begin{abstract}
Embedding layers are commonly used to map discrete symbols into continuous embedding vectors that reflect their semantic meanings. Despite their effectiveness, the number of parameters in an embedding layer increases linearly with the number of symbols and poses a critical challenge on memory and storage constraints. In this work, we propose a generic and end-to-end learnable compression framework termed differentiable product quantization (DPQ). We present two instantiations of DPQ that leverage different approximation techniques to enable differentiability in end-to-end learning. Our method can readily serve as a drop-in alternative for any existing embedding layer. Empirically, DPQ offers significant compression ratios (14-238$\times$) at negligible or no performance cost on 10 datasets across three different language tasks.~\footnote{Code at: \href{https://github.com/chentingpc/dpq\_embedding\_compression}{github.com/chentingpc/dpq\_embedding\_compression}.}
\end{abstract}
 \section{Introduction}
\label{sec:introduction}
The embedding layer is a basic neural network module which maps a discrete symbol/word into a continuous hidden vector. It is widely used in NLP related applications, including language modeling, machine translation and text classification. With large vocabulary sizes, embedding layers consume large amounts of storage and memory. For example, in the medium-sized LSTM-based model on the PTB dataset~\citep{zaremba2014recurrent}, the embedding table accounts for more than 95\% of the total number of parameters. Even with sub-words encoding~\cite{sennrich2015neural,kudo2018sentencepiece}, the size of the embedding layer is still very significant. In addition to words/sub-words models in the text domain~\citep{mikolov2013efficient,devlin2018bert}, embedding layers are also used in a wide range of applications such as knowledge graphs~\citep{bordes2013translating,socher2013reasoning} and recommender systems~\citep{koren2009matrix}, where the vocabulary sizes are even larger.

Recent efforts to reduce the size of embedding layers have been made~\citep{shu2017compressing,chen2018learning}, where the authors propose to first learn to encode symbols/words with K-way D-dimensional discrete codes (KD codes, such as 5-1-2-4 for ``cat'' and 5-1-2-3 for ``dog''), and then compose the codes to form the output symbol embedding using an embedding composition function. However, in~\cite{shu2017compressing}, the discrete codes are fixed before training and are therefore non-adaptive for downstream tasks. \cite{chen2018learning} proposes to learn discrete codes in an end-to-end fashion which leads to better task performance, but their method has two major drawbacks. Firstly, they utilize complicated embedding composition functions (e.g. recurrent networks, MLPs) to convert discrete codes into embedding vectors, which are both computationally expensive and hard to learn. Secondly, as a result, an additional distillation procedure is required in order to avoid performance drop.

In this work, we propose a simple differentiable product quantization (DPQ) framework, which is an efficient module to insert discrete codes into a neural network while maintaining differentiability. The proposal is based on the observation that the discrete codes are naturally derived through the process of quantization (product quantization by~\citet{jegou2010product} in particular), and by making the quantization process differentiable, we are able to learn the discrete codes in an end-to-end fashion. Under our framework, we present two concrete approximation techniques that allow differentiable learning. Compared to the existing methods~\cite{shu2017compressing,chen2018learning}, our framework 1) establishes a simple and general framework to generate discrete codes in neural nets in a differentiable manner, 2) enables more efficient computation and better approximations, 3) achieves better task performance \textit{and} better compression ratios, and 4) avoids the two-stage training (pre-training and distillation) as needed in ~\cite{chen2018learning}.

We conduct experiments on ten different datasets across three language tasks, with additional experiments on BERT~\cite{devlin2018bert} pre-training, by simply replacing the original embedding layer with DPQ. The results show that DPQ can learn compact discrete embeddings with higher compression ratios than existing methods, at the same time achieving the same performance as the original full embeddings. Furthermore, our results are obtained from single-stage end-to-end training. \section{Method}

\textbf{Problem setup.} An embedding function can be defined as $\mathcal{F}_{\mathcal{W}}:\mathcal{V}\rightarrow \mathbb{R}^d$, where $\mathcal{V}$ denotes the vocabulary of discrete symbols, and $\mathcal{W}\in \mathbb{R}^{n\times d}$ is the embedding table with $n=|\mathcal{V}|$. 
In standard end-to-end training, the embedding function is jointly trained with other neural net parameters to optimize a given objective.
The goal of this work is to learn a compact embedding function $\mathcal{F}_{\mathcal{W}'}$ in the same end-to-end fashion, but the number of bits used for the new parameterization $\mathcal{W}'$ is substantially smaller than the original full embedding table $\mathcal{W}$.

\textbf{Motivation.} One important discovery for constructing a compact embedding layer is to represent each symbol using a learned discrete code~\cite{shu2017compressing,chen2018learning}, and then compose the code embedding vectors to form the final symbol embedding vector via an embedding composition function. However, it is not clear how the discrete codes are generated, and what the embedding composition function should be. One could directly optimize codes as free parameters, but it is both ad-hoc and restrictive. 

The key insight in this work is that discrete codes can be naturally generated from the process of quantization (product quantization~\citep{jegou2010product} in particular) of a continuous space, and the embedding composition function to get the continuous symbol embedding from the discrete codes is simply the reverse of the quantization process. The quantization process can be specified in flexible ways, and by making this quantization process differentiable we enable single-stage end-to-end learning of discrete codes via optimizing task-specific objectives.

\subsection{Differentiable Production Quantization Framework}

The proposed differentiable production quantization (DPQ) function is a mapping between continuous spaces, i.e. $\mathcal{T}: \mathbb{R}^{d}\rightarrow\mathbb{R}^d$. In between the two continuous spaces, there is a discrete space $\{1,\cdots, K\}^D$ which can be seen as a discrete bottleneck. To transform from continuous space to discrete space and back, two important functions are used: 1) a \textit{discretization function} $\bm\phi(\cdot):\mathbb{R}^d\rightarrow \{1, \cdots, K\}^D$ that maps a continuous vector into a K-way D-dimensional discrete code, and 2) a \textit{reverse-discretization function} $\bm\rho(\cdot):\{1, \cdots, K\}^D \rightarrow\mathbb{R}^d$ that maps the discrete code into a continuous embedding vector. In other words, the general DPQ mapping is $\mathcal{T}(\cdot)=\bm\rho\circ\bm\phi(\cdot)$.

\begin{figure*}[!t]
    \centering
    \includegraphics[trim=50 30 50 70,clip,width=0.8\textwidth]{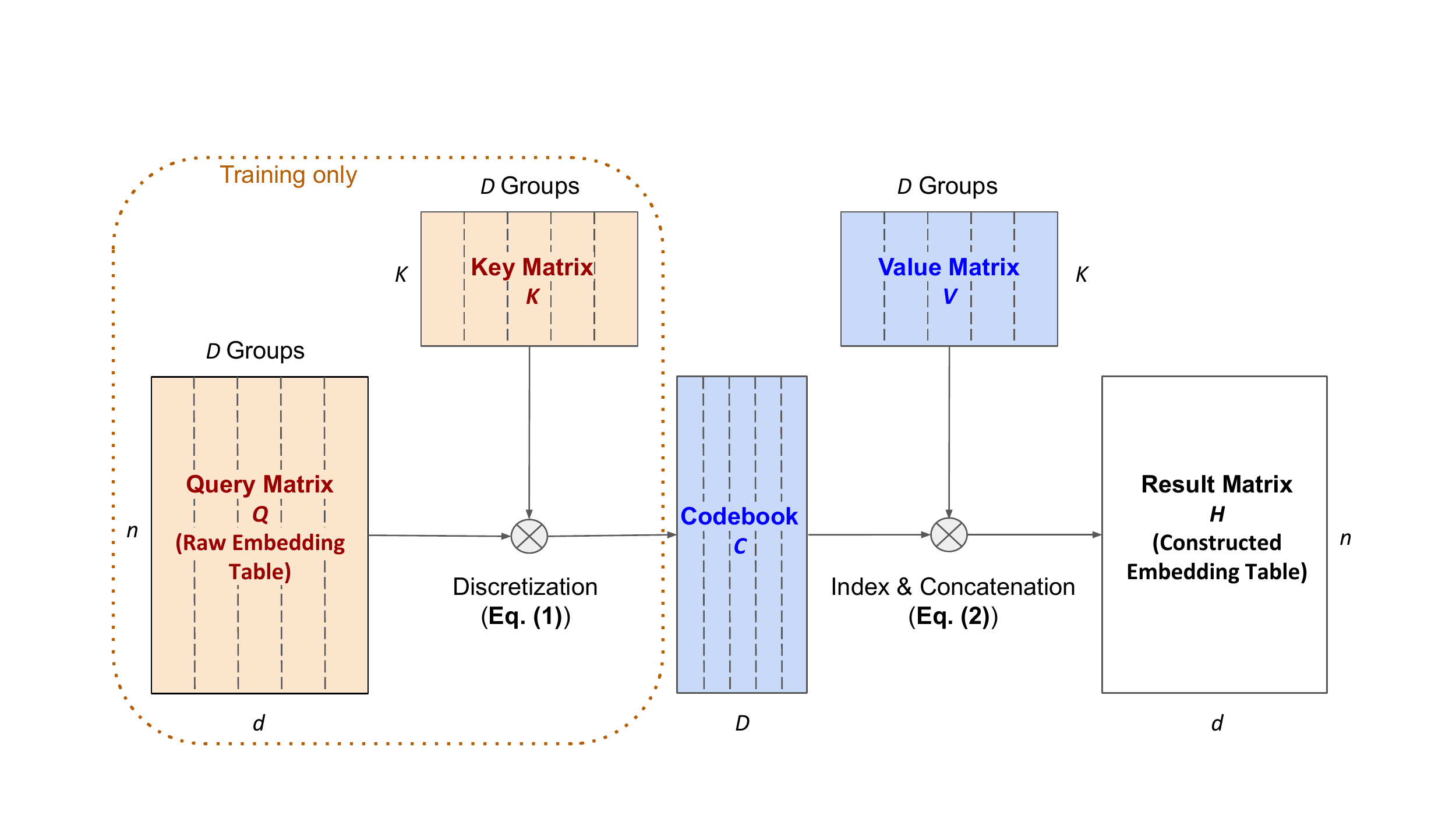}
    \caption{The DPQ embedding framework. During training, differentiable product quantization is used to approximate the raw embedding table (i.e. the query Matrix). At inference, only the codebook $\mathbf{C} \in \{1,...,K\}^{n \times D}$ and the value matrix $\mathbf{V}\in \mathbb{R}^{K\times d}$ are needed to construct the embedding table.}
    \label{fig:framework}
\end{figure*}

\paragraph{Compact embedding layer via DPQ.} In order to obtain a compact embedding layer, we first take a raw embedding and put it through DPQ function. More specifically, the raw embedding matrix can be presented as a query matrix $\mathbf{Q} \in \mathbb{R}^{n \times d}$ where the number of rows equals to the vocabulary size. The discretization function of DPQ computes discrete codes $\mathbf{C}=\bm\phi(\mathbf{Q})$ where $\mathbf{C} \in \{1, \cdots, K\}^{n\times D}$ is the discrete \textit{codebook}. To construct the final embedding table for all symbols, the reverse-discretization function of DPQ is applied, i.e. $\mathbf{H} = \bm \rho(\mathbf{C})$ where $\mathbf{H}\in \mathbb{R}^{n\times d}$ is the final symbol embedding matrix. In order to make it compact for the inference, we will discard the raw embedding matrix $\mathbf{Q}$ and only store the codebook $\mathbf{C}$ and small parameters needed in the reverse-discretization function. They are sufficient to (re)construct partial or whole embedding table. In below, we specify the discretization function $\bm\phi(\cdot)$ and reverse-discretization function $\bm\rho(\cdot)$ via product keys and values.
Figure \ref{fig:framework} illustrates the proposed framework. The proposed method can also be seen as a learned hash function of finite input into a set of discrete codes, and use lookup during the inference instead of re-compute the codes.

\paragraph{Product keys for discretization function $\bm\phi(\cdot)$.} Given the query matrix $\mathbf{Q}$, the discretization function computes the codebook $\mathbf{C}$. While it is possible to use a complicated transformation, in order to make it efficient, we simply leverage a key matrix $\mathbf{K}\in \mathbb{R}^{K\times d}$ with $K$ rows where $K$ is the number of choices for each code bit. 
In the spirit of \textit{product keys} in product quantization, we further split columns of $\mathbf{K}$ and $\mathbf{Q}$ into $D$ groups/subspace, such that $\mathbf{K}^{(j)}\in\mathbb{R}^{K\times d/D}$ and $\mathbf{Q}^{(j)}\in\mathbb{R}^{n\times d/D}$.

We can compute each of $D$ dimensional KD codes separately. The $j$-th dimension of a KD code $\mathbf{C}_i$ for the $i$-th symbol is computed as follows.
\begin{equation}
\label{eq:phi}
\mathbf{C}^{(j)}_i=\argmin_k \text{dist}\bigg(\mathbf{Q}^{(j)}_i, \mathbf{K}^{(j)}_k\bigg)
\end{equation}
The $\text{dist}(\cdot, \cdot)$ computes distance measure between two vectors, and use it to decide which discrete code to take.

\paragraph{Product values for reverse-discretization function $\bm\rho(\cdot)$.} Given the codebook $\mathbf{C}$, the reverse-discretization function computes the final continuous embedding vectors. While this can be another sophisticated transformation, we again opt for the most efficient design and employee a single value matrix $\mathbf{V}\in \mathbb{R}^{K\times d}$ as the parameter. Similarly, we leverage product keys, and split the columns of $\mathbf{V}$ into $D$ groups/subspaces the same way as $\mathbf{K}$ and $\mathbf{Q}$, i.e. $\mathbf{V}^{(j)}\in\mathbb{R}^{K\times d/D}$. We use the code in each of $D$ dimension to index the subspace in $\mathbf{V}$, and concatenate the results to form the final embedding vector as follows.
\begin{equation}
\label{eq:rho}
    \mathbf{H}_i = [\mathbf{V}^{(1)}_{\bm c^{(1)}_i}, \cdots, \mathbf{V}^{(j)}_{\bm c^{(j)}_i},\cdots, \mathbf{V}^{(D)}_{\bm c^{(D)}_i}]
\end{equation}
The inference algorithm for the $i$-th token is given in Algorithm~\ref{alg:infer}.

\begin{algorithm}[!t]
\begin{algorithmic}
\REQUIRE $\mathbf{V} \in \mathbb{R}^{K\times D \times(d/D)}$, $\mathbf{C} \in \{1, ..., K\}^{n\times D}$
\FOR{$j \in \{1,...,D\}$}
  \STATE $\bm h^{(j)}_i = \mathbf{V}^{(j)}_{\mathbf{C}^{(j)}_i}$
\ENDFOR
\STATE \textbf{return} $\text{concatenate}(\bm h^{(1)}_i, \bm h^{(2)}_i, ..., \bm h^{(D)}_i)$
\end{algorithmic}
\caption{\label{alg:infer}Inference of embedding for $i$-th token.}
\end{algorithm}

\textbf{Inference complexity.} Since only indexing and concatenation (Eq. \ref{eq:rho}) are used during inference, both the extra computation complexity and memory footprint are usually negligible compared to the regular full embedding (which directly indexes an embedding table).

\textbf{Storage complexity.} Assuming the default 32-bit floating point is used, the original full embedding table requires $32nd$ bits. As for DPQ embedding, we only need to store the codebook and the value matrix: 1) codebook $\mathbf{C}$ requires $nD\log_2K$ bits, which is the only thing that depends on vocabulary size $n$, and 2) value matrix $\mathbf{V}$ requires $32Kd$ bits\footnote{$32Kd/D$ bits if we share the weights among $D$ groups/subspaces.}, which does not explicitly depend on $n$ and is negligible when $n$ is large. Since typically $nD\log_2K \ll 32nd$, the DPQ embedding is more compact than the full embedding.

\textbf{Expressiveness.} DPQ achieves compactness by introducing sparsity into the embedding matrix in two axis: (1) the product keys/values, and (2) top-1 selection in each group/subspace. DPQ is able to achieve compactness without decreasing the rank of the resulting embedding matrix.

\begin{proposition}\label{theorem}
The DPQ embedding matrix $\mathbf{H}$ is full rank given the following constraints are satisfied.
\begin{itemize}
\itemsep-0.2em
    \item[1)] One-hot encoded $\mathbf{C} \in \{1,...,K\}^{n \times D}$, denoted as $\mathbf{B}\in\{0,1\}^{n\times KD}$, is full-rank.
    \item[2)] Sub-matrices of splitted $\mathbf{V}$, i.e. $\mathbf{V}^{(j)} \in \mathbb{R}^{K\times d/D}, \forall j$, are all full-rank.
    \item[3)] $KD \ge d$.
\end{itemize}
\end{proposition}
The proof is given in the appendix \ref{app:proof}. 
Note that the above conditions are easy to satisfy while being compact than full embedding, since it is easy to satisfy $nD\log_2K \ll 32nd$ with $KD \ge d$ in practice.

So far we have not specified designs of discretization functions such as the distance function in Eq \ref{eq:phi}. More importantly, how can we compute gradients through the $\argmin$ function in Eq. \ref{eq:phi}? While there could be many instantiations with different design choices, below we introduce two instantiations that use two different approximation schemes for DPQ.

\begin{figure*}[t]
\centering
    \begin{subfigure}[b]{0.55\textwidth}
        \centering
        \includegraphics[trim=20 270 370 20,clip,scale=0.77]{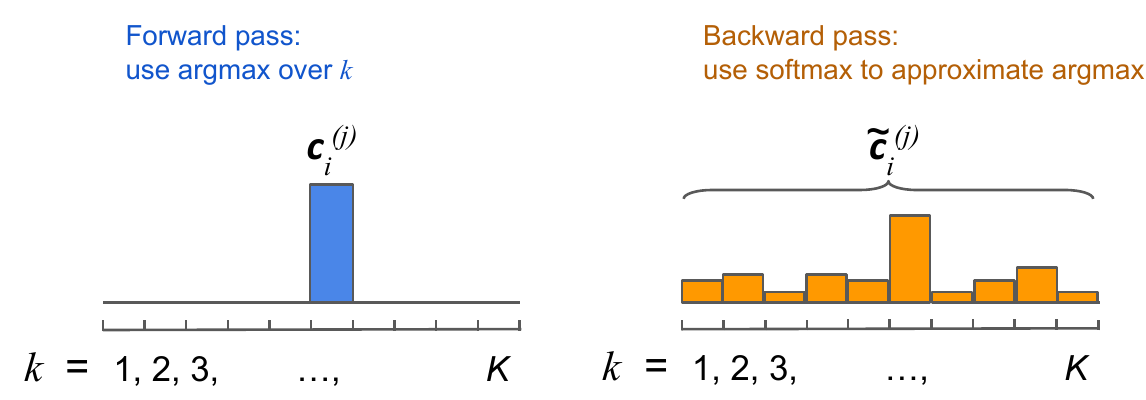}
        \caption{Softmax-based (DPQ-SX)}
    \end{subfigure}
    \begin{subfigure}[b]{0.44\textwidth}
        \centering
        \includegraphics[trim=20 220 440 50,clip,scale=0.65]{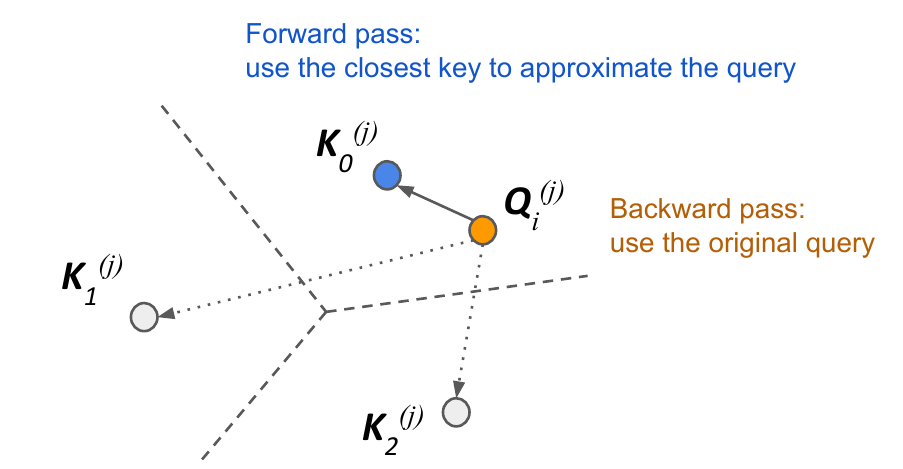}
        \caption{Centroid-based (DPQ-VQ)}
    \end{subfigure}
\caption{\label{fig:approximations}Illustration of two types of approximation to enable differentiability in DPQ.}
\end{figure*}

\begin{table*}[t]
\begin{center}
\begin{small}
\begin{tabular}{lllll}
\toprule
\bf Method &\bf Dist. Metric &\bf Key/Value matrices  &\bf Train &\bf Inference
\\ \midrule
DPQ-SX & Dot product and more & Not tied, allows different sizes & Efficient & Efficient \\ \midrule
DPQ-VQ & Euclidean only & Tied & More efficient & Efficient \\ \bottomrule
\end{tabular}
\end{small}
\caption{\label{tab:sxvq}Summary of differences between VQ and SX. DPQ-SX allows more flexibility in distance metrics and whether to tie the key and value metrices. DPQ-VQ is more efficient during training and therefore is more scalable to larger $K,D$ values.}
\end{center}
\end{table*}

\subsection{Softmax-based Approximation}
The first instantiation of DPQ (named DPQ-SX) approximates the non-differentiable $\argmax$ operation with a differentiable softmax function. 
To do so, we first specify the distance function in Eq. \ref{eq:phi} with a softmax function as follows.
\begin{equation}
\mathbf{C}^{(j)}_i = \argmax_k  \frac{\exp(\langle\mathbf{Q}^{(j)}_i, \mathbf{K}^{(j)}_k\rangle)}{\sum_{k'} \exp(\langle\mathbf{Q}^{(j)}_i, \mathbf{K}^{(j)}_{k'}\rangle)}
\end{equation}
where $\langle\cdot,\cdot\rangle$ denotes dot product of two vectors (alternatively, other metrics such as Euclidean distance, cosine distance can also be used). To approximate the $\argmax$, similar to \citep{jang2016categorical,chen2018learning}, we relax the softmax function with temperature $\tau$:
\begin{equation}
\label{eq:dist_sx}
\tilde{\mathbf{C}}^{(j)}_i =\exp(\langle\mathbf{Q}^{(j)}_i, \mathbf{K}^{(j)}_k\rangle/\tau)/ Z
\end{equation}
where $Z=\sum_{k'} \exp(\langle\mathbf{Q}^{(j)}_i, \mathbf{K}^{(j)}_{k'}\rangle/\tau)$. Note that now $\tilde{\mathbf{C}}^{(j)}_i \in \Delta^{K}$ is a probabilistic vector (i.e. soft one-hot vector) instead of an integer $\mathbf{C}^{(j)}_i$. And $\text{one\_hot}(\mathbf{C}^{(j)}_i)\approx \tilde{\mathbf{C}}^{(j)}_i$, or $\mathbf{C}^{(j)}_i=\argmax \tilde{\mathbf{C}}^{(j)}_i$. With a one-hot code relaxed into soft one-hot vector, we can replace index operation $\mathbf{V}^{(j)}_{\tilde{\mathbf{C}}^{(j)}_i}$ with dot product to compute the output embedding vector, i.e. $\mathbf{H}^{(j)}_i = \tilde{\mathbf{C}}^{(j)}_i \mathbf{V}^{(j)}$.

The softmax approximated computation defined above is fully differentiable when $\tau \neq 0$. However, to compute discrete codes during the forward pass, we have to set $\tau\rightarrow 0$, which turns the softmax function into a spike concentrated on the $\mathbf{C}^{(j)}_i$-th dimension. This is equivalent to the $\argmax$ operation which does not have gradient.

To enable a pseudo gradient while still be able to output discrete codes, we use a different temperatures during forward and backward pass, i.e. set $\tau= 0$ in forward pass, and $\tau= 1$ in the backward pass. So the final DPQ function can be expressed as follows.
\begin{equation}
\mathbf{H}_i= \mathcal{T}(\mathbf{Q}_i|\tau=1) - \text{sg}\bigg(\mathcal{T}(\mathbf{Q}_i|\tau=1) -\mathcal{T}(\mathbf{Q}_i|\tau=0)\bigg)
\end{equation}
Where $\text{sg}$ is the \textit{stop gradient} operator, which is identity function in forward pass, but drops gradient for variables inside it during the backward pass. And $\mathcal{T}(\cdot)=\bm\rho\circ\bm\phi(\cdot)$ is the DPQ mapping defined by Eq.~\ref{eq:phi},~\ref{eq:rho} with distance function specified by Eq~\ref{eq:dist_sx}.

\begin{table*}[!t]
\small
\centering
\begin{tabular}{llllp{5.8cm}}
\toprule
\bf Task &\bf Dataset &\bf Vocab Size  &\bf Tokenization &\bf Base Model
\\ \midrule
\multirow{2}{*}{LM} &PTB &10,000 &\multirow{2}{*}{Words} &\multirow{2}{5.2cm}{LSTM-based models from~\cite{zaremba2014recurrent}, three model sizes} 
\\ &Wikitext-2 &33,278 & &
\\ \midrule
\multirow{3}{*}{NMT} &IWSLT15 (En-Vi) &17,191 &\multirow{2}{*}{Words} &\multirow{2}{5.2cm}{Seq2seq-based model from~\cite{luong17}}
\\ &IWSLT15 (Vi-En) &7,709 & &
\\ \cmidrule{2-5} 
&WMT19 (En-De) &32,000 &Sub-words & Transformer Base in \cite{vaswani2017attention}
\\ \midrule
\multirow{5}{*}{TextC} &AG News &69,322 &\multirow{5}{*}{Words} &\multirow{5}{5.2cm}{One hidden layer after mean pooling of word vectors, similar to fastText from~\cite{joulin2017bag}}
\\ &Yahoo! Ans. &477,522 & &
\\ &DBpedia &612,530 & &
\\ &Yelp P &246,739 & &
\\ &Yelp F &268,414 & & \\
\bottomrule
\end{tabular}
\caption{\label{tab:datasets_and_models}Datasets and models used in our experiments. More details in Appendix \ref{app:datasets_and_models}.}
\end{table*}

\subsection{Centroid-based Approximation}
The second instantiation of DPQ (named DPQ-VQ) uses a centroid-based approximation, which directly pass the gradient straight-through~\citep{bengio2013estimating} a small set of centroids. In order to do so, we need to put $\mathbf{Q}, \mathbf{K}, \mathbf{V}$ into the same space.

First, we treat rows in key matrix $\mathbf{K}$ as centroids, and use them to approximate query matrix $\mathbf{Q}$. The approximation is based on the Euclidean distance as follows.
\begin{equation}
\mathbf{C}^{(j)}_i = \argmin_k \|\mathbf{Q}^{(j)}_i - \mathbf{K}^{(j)}_k\|^2
\end{equation}
Secondly, we tie the key and value matrices, i.e. $\mathbf{V} = \mathbf{K}$, so that we can pass the gradient through.

We still have the non-differentiable $\argmin$ operation, and the input query $\mathbf{Q}^{(j)}_i$ are different from selected output centroid $\mathbf{V}^{(j)}_{\mathbf{C}^{(j)}_i}$. However, since they are in the same space, it allows us to directly pass the gradient straight-through as follows.
\begin{equation}
\label{eq:vq_sg}
\mathbf{H}_i = \mathbf{Q}_i - \text{sg}(\mathbf{Q}_i - \mathcal{T}(\mathbf{Q}_i))
\end{equation}
Where $\text{sg}$ is again the \textit{stop gradient} operation. During the forward pass, the selected centroid is emitted, but during the backward pass, the gradient is pass to the query directly. This provides a way to compute discrete codes in the forward pass (which are the indexes of the centroids), and update the query matrix during the backward pass.

However, it is worth noting that the Eq. \ref{eq:vq_sg} only approximates gradient for query matrix, but does not updates the centroids, i.e. the tied key/value matrix. Similar to~\cite{van2017neural}, we add a regularization term: $\mathcal{L}_{reg} = \sum_i\|\mathcal{T}(\mathbf{Q}_i) - \text{sg}(\mathbf{Q}_i)\|^2$, which makes entries of the key/value matrix arithmetic mean of their members. Alternatively, one can also use Exponential Moving Average~\citep{kaiser2018fast} to update the centroids.

\textbf{A comparison between DPQ-SX and DPQ-VQ.} DPQ-VQ and DPQ-SX only differ during training. They are different in how they approximate the gradient for the non-differentiable $\argmin$ function: DPQ-SX approximates the one-hot vector with softmax, while DPQ-VQ approximates the continuous vector using a set of centroids. Figure \ref{fig:approximations} illustrates this difference. This suggests that when there is a large gap between one-hot and probabilistic vectors (large $K$), DPQ-SX approximation could be poor; and when there is a large gap between the continuous vector and the selected centroid (large subspace dimension, i.e. small $D$), DPQ-VQ could have a big approximation error.

Table \ref{tab:sxvq} summarizes the comparisons between DPQ-SX and DPQ-VQ. DPQ-SX is more flexible as it does not constrain the distance metric, nor does it tie the key/value matrices as in DPQ-VQ. Thus one could use different sizes of key and value matrices. Regarding to the computational cost during training, DPQ-SX back-propagates through the whole distribution of $K$ choices, while DPQ-VQ only back-propagates through the nearest centroid, making it more scalable (to large $K$, $D$, and batch sizes).

\subsection{Implementation Details}

\paragraph{Distance normalization.} Training with straight-through estimator can be unstable as the gradient is approximated. To mitigate this problem, we apply batch normalization~\cite{ioffe2015batch} for the distance measure in DPQ along the K-dimension, i.e. each centroid will have a normalized distance distribution over batch samples.

\paragraph{Subspace-sharing.} To further reduce parameters, one can share parameters among the $D$ groups in the Key/Value Matrices, i.e. constraining $\mathbf{K}^{(j)}=\mathbf{K}^{(j')}$ and $\mathbf{V}^{(j)}=\mathbf{V}^{(j')}, \forall j, j'$. For simplicity we refer to this as "subspace-sharing". We search over both options and utilize subspace-sharing if there is no performance drop. \section{Experiments}

We conduct experiments on ten datasets across three tasks: language modeling (LM), neural machine translation (NMT) and text classification (TextC)~\cite{zhang2015character}.
We adopt existing architectures for these tasks as base models and only replace the input embedding layer with DPQ embeddings. The details of datasets and base models are summarized in Table \ref{tab:datasets_and_models}.

We evaluate the models using two metrics: task performance and compression ratio. Task performance metrics are perplexity scores for LM tasks, BLEU scores for NMT tasks, and accuracy in TextC tasks. Compression ratios for the embedding layer is computed as follows:
$$
\text{CR} = \frac{\text{\# of bits used in the full embedding table}}{\text{\# of bits used in compressed model for inference}}
$$
For DPQ in particular, this can be computed as $\text{CR} = \frac{32nd}{nD\log_2 K + 32Kd}$. Further compression can be achieved with `subspace-sharing', with which we have $\text{CR} = \frac{32nd}{nD\log_2 K + 32Kd/D}$. In this work, we focus on the embedding table in the encoder side, so we keep the decoder embedding layer (i.e. output softmax layer) as is.

\begin{table*}[t]
\small
\centering
\begin{tabular}{lllccccc}
\toprule
\textbf{Task} & \textbf{Metric} & \textbf{Dataset} & \textbf{Baseline} & 
\textbf{DPQ-SX} & \textbf{(Compr. Ratio$\uparrow$)} & \textbf{DPQ-VQ} & \textbf{(Compr. Ratio$\uparrow$)} \\
\midrule
\multirow{2}{*}{LM} & \multirow{2}{*}{PPL$\downarrow$} & PTB & 83.4 & \textbf{83.2} & \textbf{(163.2)} & 83.3 & (58.7) \\
& & Wikitext-2 & 95.6 & \textbf{95.0} & \textbf{(59.3)} & 95.9 & (95.3) \\
\midrule
\multirow{3}{*}{NMT} & \multirow{3}{*}{BLEU$\uparrow$}
& IWSLT15 (En-Vi) & \textbf{25.4} & 25.3 & (86.2) & 25.3 & (16.1) \\
& & IWSLT15 (Vi-En) & 23.0 & \textbf{23.1} & \textbf{(72.0)} & 22.5 & (14.1) \\
& & WMT19 (En-De) & \textbf{38.8} & \textbf{38.8} & \textbf{(18.0)} & 38.7 & (18.2) \\
\midrule
\multirow{5}{*}{TextC} & \multirow{5}{*}{Acc(\%)$\uparrow$} 
& AG News &\textbf{92.6} & 92.5 & (19.3) & 92.6 & (24.0) \\
& & Yahoo! Ans. & 69.4 & \textbf{69.6} & \textbf{(48.2)} & 69.2 & (19.2) \\
& & DBpedia & 98.1 & 98.1 & (24.1) & \textbf{98.1} & \textbf{(38.5)} \\
& & Yelp P & 93.9 & \textbf{94.2} & \textbf{(38.5)} & 93.9 & (24.0) \\
& & Yelp F & \textbf{60.3} & 60.1 & (48.2) & 60.2 & (24.1) \\
\bottomrule
\end{tabular}
\caption{\label{tab:all_tasks}Comparisons of DPQ variants vs. the full embedding baseline on ten datasets across three tasks. We use $\downarrow$ to denote the lower the better, in contrast, $\uparrow$ means the higher the better.}
\end{table*}

\subsection{Compression Ratios and Task Performance Against Baselines}
\begin{table}[t]
\small
\centering
\begin{tabular}{lcccccc}
\toprule
& \multicolumn{2}{c}{\textbf{Small}} & \multicolumn{2}{c}{\textbf{Medium}} & \multicolumn{2}{c}{\textbf{Large}} \\
\textbf{Method} & \textbf{PPL$\downarrow$} & \textbf{CR$\uparrow$} & \textbf{PPL$\downarrow$} & \textbf{CR$\uparrow$} & \textbf{PPL$\downarrow$} & \textbf{CR$\uparrow$} \\
\midrule
Full & 114.5 & 1 & 83.4 & 1 & 78.7 & 1 \\
\midrule
Shu$'$17& 108.0 & 4.8 & 84.9 & 12.5 & 80.7 & 18.5 \\
Chen$'$18& 108.5 & 4.8 & 89.0 & 12.5 & 86.4 & 18.5 \\
Chen$'$18+& 107.8 & 4.8 & 83.1 & 12.5 & \textbf{77.7} & 18.5 \\
\midrule
{DPQ-SX}& \textbf{105.8} & \textbf{85.5} & \textbf{82.0} & \textbf{82.9} & 78.5 & \textbf{238.3} \\
{DPQ-VQ} & 106.5 & 51.1 & 83.3 & 58.7 & 79.5 & \textbf{238.3} \\
\bottomrule
\end{tabular}
\caption{\label{tab:baselines}Comparison of DPQ against recently proposed embedding compression techniques on the PTB LM task (LSTMs with three model sizes are studied). Metrics are perplexity (PPL) and compression ratio (CR).}
\end{table}

\paragraph{Comparison with full embedding baseline on ten datasets.} Table \ref{tab:all_tasks} summarizes the task performance and compression ratios of DPQ-SX and DPQ-VQ against baseline models that use the regular full embeddings\footnote{For LM, results are from the medium-sized LSTM model.}. In each task/dataset, we report results from a configuration that gives as good task performance as the baseline (or as good as possible, if it does not match the baseline) while providing the largest compression ratio. In all tasks, both DPQ-SX and DPQ-VQ can achieve comparable or better task performance while providing a compression ratio from $14\times$ to $163\times$. In 6 out of 10 datasets, DPQ-SX performs strictly better than DPQ-VQ in both metrics. \textit{Remarkably, DPQ is able to further compress the already-compact sub-word representations used in WMT19 (En-De). This shows great potential of DPQ to learn very compact embedding layers.}

\textbf{Comparison with more baselines on LM.} Here we compare DPQ against the recently proposed embedding compression methods. Shu$'$17 from~\cite{shu2017compressing}: a three-step procedure where one firstly trains a full model, secondly learns discrete codes to reconstruct the pre-trained embedding layer, and finally fixes the discrete codes and trains the model again; Chen$'$18 from~\cite{chen2018learning}: end-to-end training without distillation guidance from a pre-trained embedding table; Chen$'$18+ from~\cite{chen2018learning}: an end-to-end training with an additional distillation procedure that uses a pre-trained embedding as guidance during training. Table \ref{tab:baselines} shows the comparison between DPQ and the above methods on the PTB language modeling task using LSTMs with three different model sizes. We find that 1) other than DPQ, only Chen$'$18+ which requires an extra distillation procedure is able to achieve similar perplexity scores as the full embedding baseline; 2) DPQ variants (particularly DPQ-SX) are able to obtain extremely competitive perplexity scores in all cases, while offering compression ratios that are an order of magnitude larger than all the other alternatives (and is trained end-to-end in a single stage).

Table \ref{tab:more_compare_lm} shows comparisons on the PTB language modeling task (using medium-sized LSTM) with broader set of baselines (including methods that are not based on discrete codes). We find that 1) traditional compression techniques, such as scalar and product quantization as well as low-rank factorization, typically degenerates the performance significantly in order to achieve good compression ratios compared to discrete code learning-based methods~\citep{shu2017compressing,chen2018learning}; 2) DPQ can largely improve the compression ratio while achieving similar or better task performance (perplexity in this case).

\begin{table}[!t]
    \small
    \centering
    \begin{tabular}{lcc}
    \toprule
    \textbf{Method} & \textbf{PPL$\downarrow$} & \textbf{Compr. Ratio$\uparrow$} \\
    \midrule
    Full & 83.4 & 1.0\\
    \midrule
    Scalar quantization (8 bits) &	84.1 &	4.0 \\
    Scalar quantization (6 bits) &	87.7 &	5.3 \\
    Scalar quantization (4 bits) &	92.9 &	8.3 \\
    Product quantization(64x325) &	84.0 &	8.3 \\
    Product quantization(128x325) &	83.7 &	6.7 \\
    Product quantization(256x325) &	83.7 &	5.3 \\
    Low-rank (5X) &	84.8 &	5.0 \\
    Low-rank (10X) & 85.5 &	10.2 \\
    \midrule
    Ours (DPQ-VQ) &	83.3 &	58.7 \\
    Ours (DPQ-SX) &	\textbf{82.0} &	\textbf{82.9} \\
    \bottomrule
    \end{tabular}   
    \caption{\label{tab:more_compare_lm}Comparison of DPQ against traditional embedding compression techniques on the PTB LM task (medium-sized LSTM).}
\end{table}

\begin{table*}[t]
    \small
    \centering
    \begin{tabular}{cccccc}
    \toprule
	\textbf{Dataset} & \textbf{AG News} & \textbf{Yahoo!} & \textbf{DBPedia} & \textbf{Yelp P} & \textbf{Yelp F} \\
	\midrule
    Full & 92.6 (1.0) & 69.4 (1.0) & 98.1 (1.0) & 93.9 (1.0) & 60.3 (1.0) \\
    \midrule
    Low-rank(10$\times$) & 91.4 (10.4) & 69.5 (10.2) & 97.7 (10.3) & 92.4 (10.4) & 57.8 (10.3) \\
    Low-rank(20$\times$) & 91.5 (21.4) & 69.1 (21.5) & 97.9 (21.3) & 92.4 (21.5) & 57.3 (21.4) \\
    \cite{chen2018learning} & 91.6 (53.3) & 69.5 (31.7) & 98.0 (48.4) & 93.1 (48.6) & 59.0 (54.4) \\
    \midrule
    DPQ-VQ & \textbf{92.6 (24.0)} & 69.2 (19.2) & \textbf{98.1  (38.5)} & 93.9 (24.0)  & 60.2 (24.1) \\
    DPQ-SX & 92.5 (19.3) & \textbf{69.6 (48.2)} & 98.1 (24.1) & \textbf{94.2 (38.5)}	& \textbf{60.1 (48.2)} \\
    \bottomrule
    \end{tabular}
    \caption{\label{tab:more_compare_tc}Performance comparison on text classification task. The accuracy and compression ratios (in parenthesis) are shown below. The proposed method (DPQ) usually achieves better accuracies than baselines, at the same time providing better compression ratios.}
\end{table*}

\begin{table*}[t]
\small
\centering
\begin{tabular}{lccccccc}
\toprule
\textbf{Embeddings} & \textbf{CR} & \textbf{Squad 1.1} & \textbf{Squad 2.0} & 
\textbf{CoLA} & \textbf{MNLI} & \textbf{MRPC} & \textbf{XNLI} \\
\midrule
Full & 1.0 & 90.1$\pm$ 0.1 /83.1$\pm$ 0.3 & 78.8$\pm$0.6/75.5$\pm$0.6 & 80.6$\pm$0.7 & 84.3$\pm$0.1 & 85.9$\pm$0.5 & 53.5$\pm$0.4 \\
DPQ-SX & 37.0 & 90.0$\pm$0.1 /83.0$\pm$0.2 & 78.7$\pm$0.5/75.4$\pm$0.5 & 80.2$\pm$0.6 & 83.7$\pm$0.2 & 85.1$\pm$0.6 & 53.4$\pm$0.1 \\
\bottomrule
\end{tabular}
\caption{\label{tab:bert}Effect of using DPQ on BERT. DPQ gives a compression ratio of $37\times$ on the embedding table while the model's performance on downstream tasks remains competitive.}
\end{table*}

\begin{table}[t]
    \small
    \centering
    \begin{tabular}{l|c|c}
        \toprule
        \textbf{Method} & \textbf{BLEU$\uparrow$} &  \textbf{Compr. Ratio$\uparrow$}\\
        \midrule
        Full & 38.8 & 1 \\
        \midrule
PQ (K=128, D=64) & 28.9 & 31.9 \\
        PQ (K=32, ~~D=128) & 35.4 & 25.0 \\
        PQ (K=128, D=128) & 35.7 & 17.0 \\
        PQ (K=32, ~~D=256) & 36.9 & 12.6 \\
        PQ (K=128, D=256) & 37.8 & 8.8 \\
        \midrule
        DPQ-VQ (K=32, D=128) & \textbf{38.7} & \textbf{17.0} \\
        DPQ-SX (K=32, D=128) & \textbf{38.8} & \textbf{17.0} \\
        \bottomrule
    \end{tabular}
    \caption{\label{tab:more_compare_nmt}Comparison of end-to-end DPQ against using PQ to reconstruct the embedding table after the model is trained, for the NMT task on WMT19 (En-DE).}
\end{table}

\textbf{Compared with more baselines on NMT.} We compare DPQ with product quantization, where we first train the full embedding model, and then learn discrete codes to reconstruct the learned full embedding table. Finally, the reconstructed embedding table replaces the original embedding table for inference. In our experiment, we use auto-encoder and DPQ (with different $K$ and $D$) to learn to reconstruct the full embedding table.

Table \ref{tab:more_compare_nmt} shows comparisons between DPQ and PQ baselines on WMT19 (En-De) with Transformer~\citep{vaswani2017attention}. We can see that PQ degenerates the performance significantly. This is expected as small approximation errors in the embedding layer accumulate and can be amplified as the errors propagate through the deep neural nets, finally leading to large errors in the output space. End-to-end training does not have this problem as the whole system is jointly learned, so the networks can account for small approximation errors in the early layers.

\textbf{Comparison with more baselines on TextC.} Table \ref{tab:more_compare_tc} provides performance comparisons on text classification tasks. We found that the proposed method (DPQ) usually achieves better accuracies than baselines, at the same time providing better compression ratios.

\subsection{Applying DPQ to BERT}
To further test DPQ, we replace the embedding layer in BERT with our DPQ for both pre-training and fine-tuning. We do not perform hyper-parameter search for DPQ, but simply use the best configuration from our experiments on WMT19 EnDe using Transformer, i.e. we use DPQ-SX with no subspace-sharing with $K=32,D=128$. For both pre-training and fine-tuning, we use the same exact configurations and hyper-parameters as in original BERT-base in~\citep{devlin2018bert}. Table \ref{tab:bert} shows that DPQ performs on par with full embedding in most of the downstream tasks, while giving a compression ratio of $37\times$ on the embedding table.

\begin{figure*}[t]
\centering
    \begin{subfigure}[b]{0.45\textwidth}
        \includegraphics[width=\textwidth]{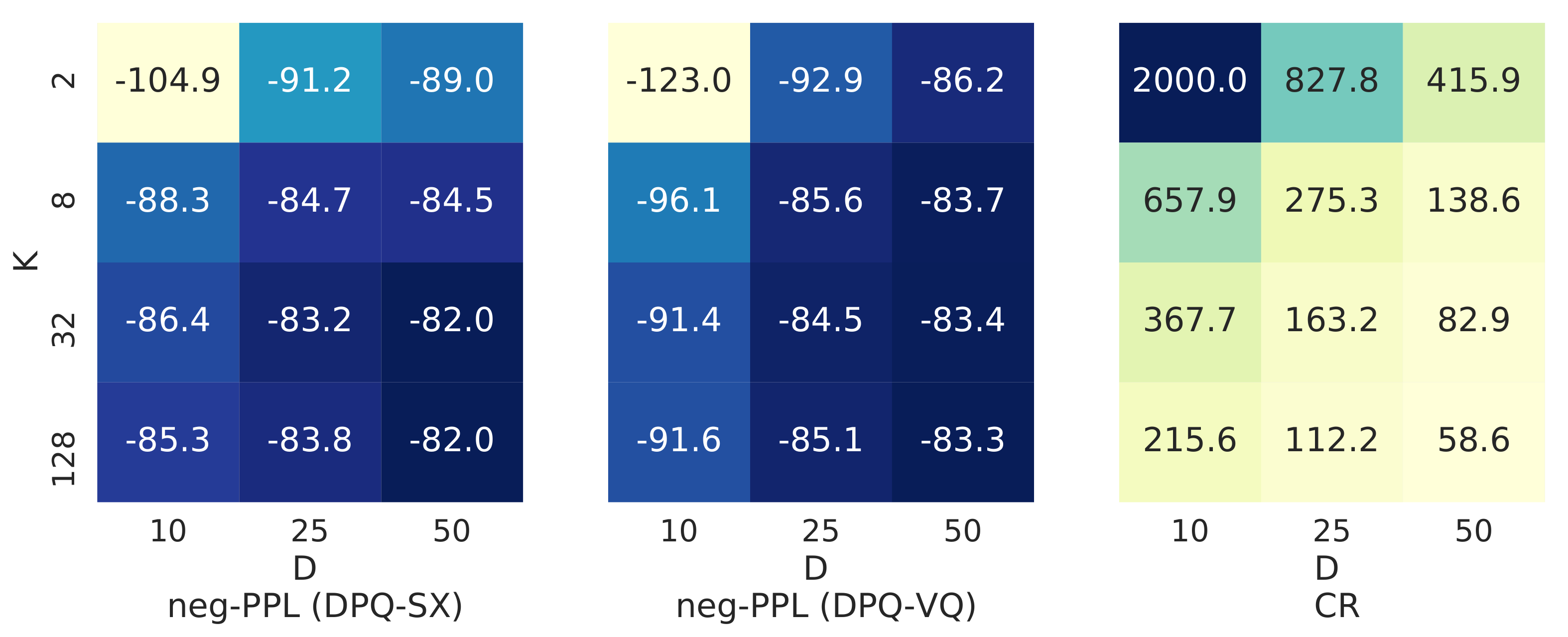}
        \caption{PTB}
    \end{subfigure}
    ~~~~
    \begin{subfigure}[b]{0.45\textwidth}
        \includegraphics[width=\textwidth]{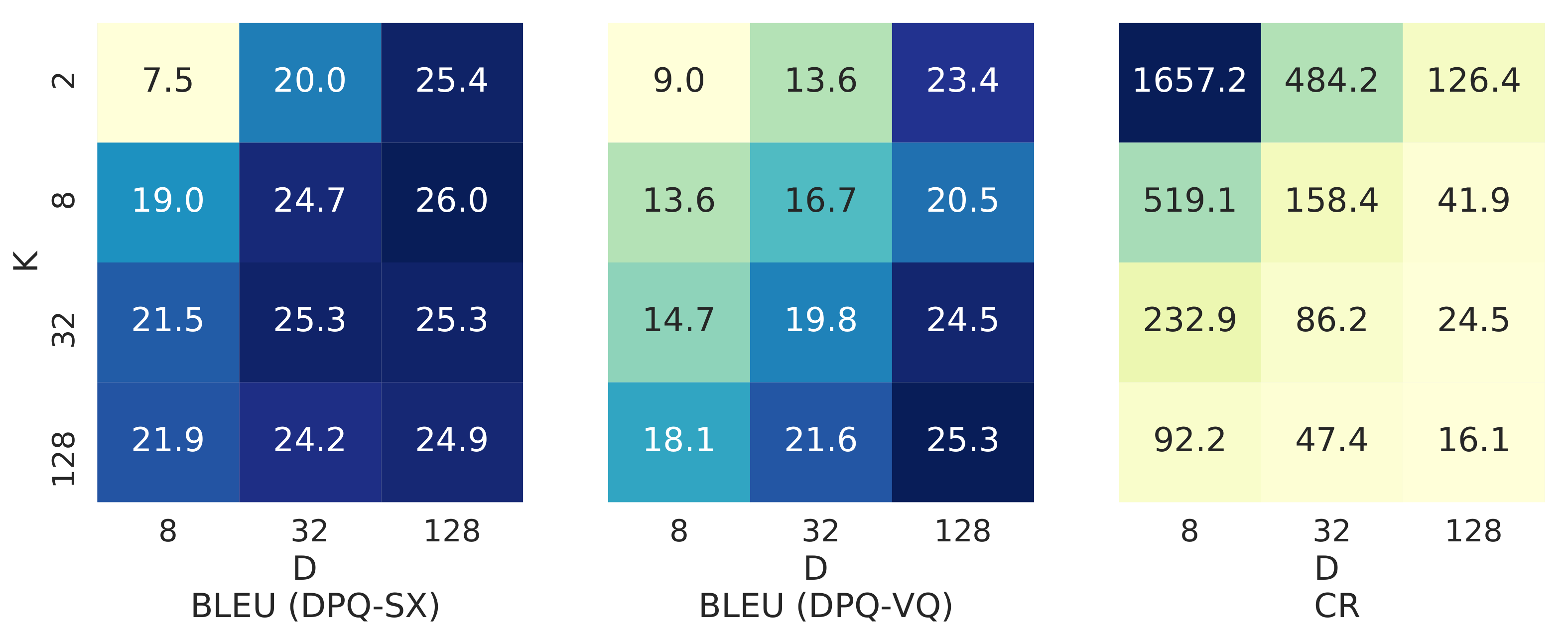}
        \caption{IWSLT15 (En-Vi)}
    \end{subfigure}
\vspace{-0.5em}
\caption{\label{fig:heatmap}Heat-maps of task performance and compression ratio for various $K$ and $D$ values. Darker is better. Key observations are: 1) increasing $K$ or $D$ typically improves the task performance at the expense of lower CRs; 2) the combination of a small $K$ and a large $D$ is better than the other way round.}
\vspace{-0.5em}
\end{figure*}

\begin{figure}[t]
\centering
    \begin{subfigure}{0.238\textwidth}
        \includegraphics[width=\textwidth]{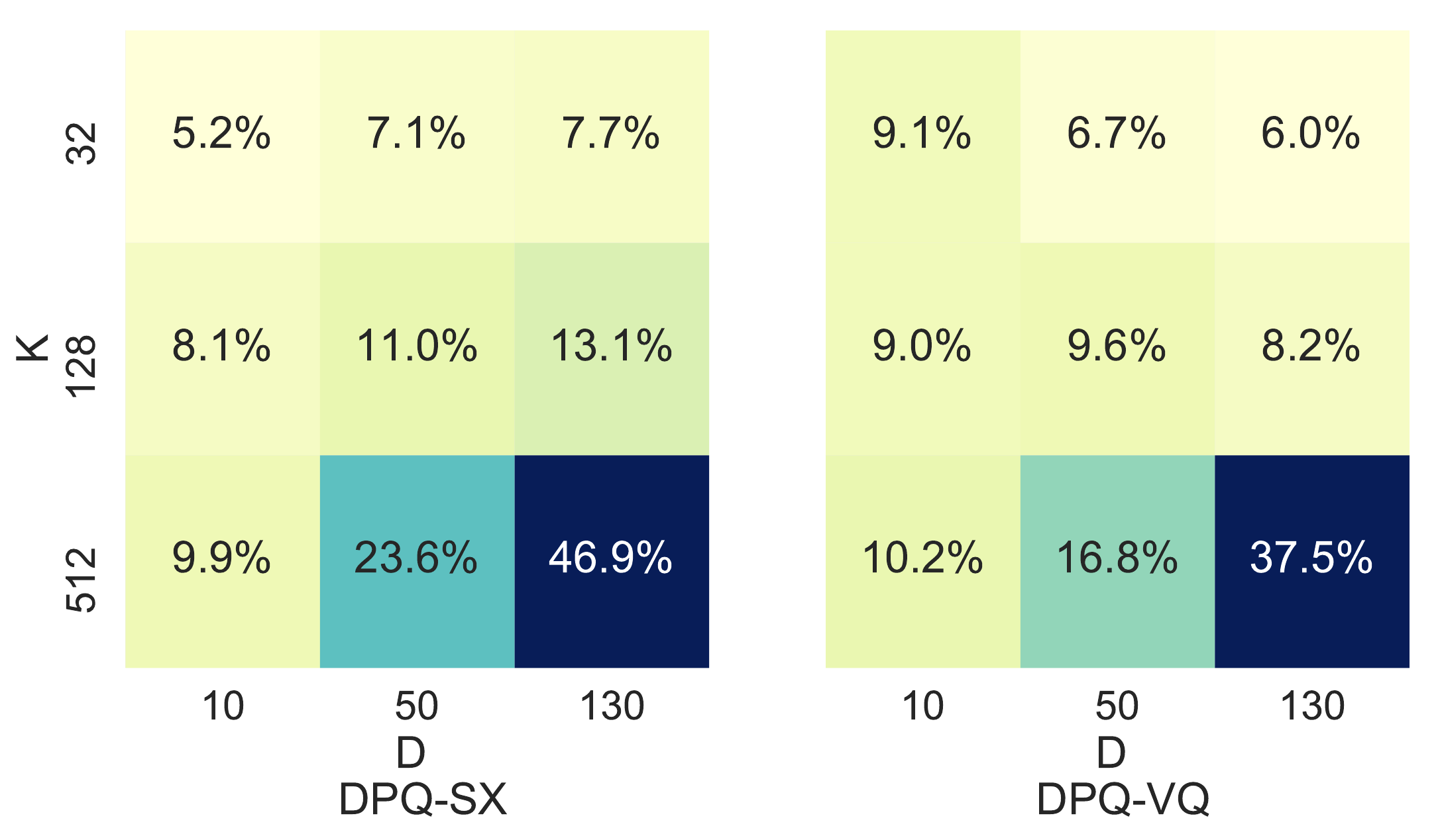}
        \caption{Extra training time used.}
    \end{subfigure}
    \begin{subfigure}{0.238\textwidth}
        \includegraphics[width=\textwidth]{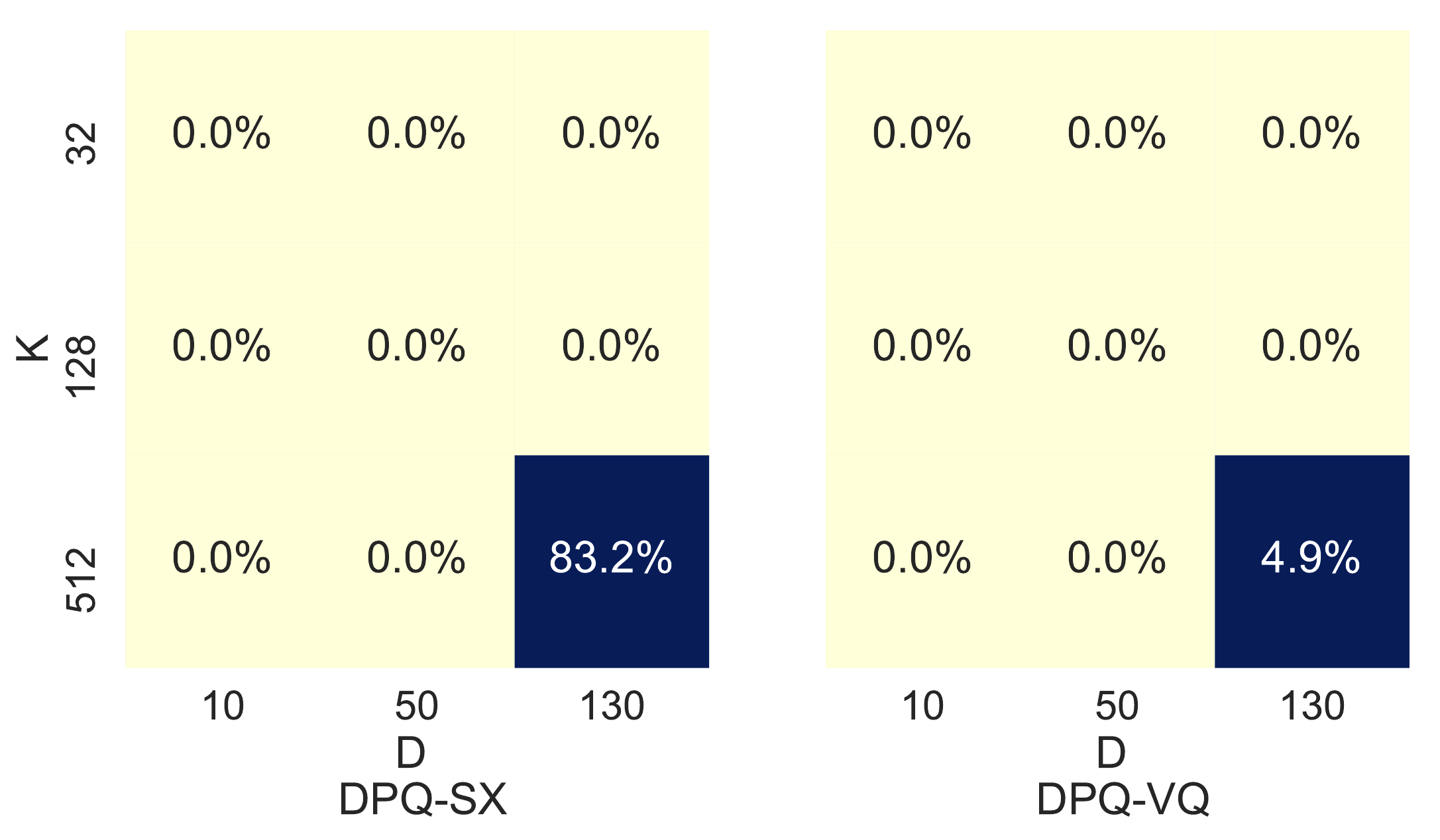}
        \caption{Extra training memory used.}
    \end{subfigure}
\caption{\label{fig:train_efficiency} Extra training cost incurred by DPQ, measured on a medium sized LSTM for LM trained on Tesla-V100 GPUs. For most $K$ and $D$ values, the extra training time is within 10\%, and the extra memory usage is zero.}
\vspace{-1em}
\end{figure}

\subsection{Effects of $K$ and $D$}
\label{sec:k_and_d}

Among key hyper-parameters of DPQ are the code size: $K$ the number of centroids per dimension and $D$ the code length. Figure \ref{fig:heatmap} shows the task performance and compression ratios for different $K$ and $D$ values on PTB and IWSLT15 (En-Vi). Firstly, we observe that the combination of a small $K$ and a large $D$ is a better configuration than the other way round. For example, in IWSLT15 (En-Vi), ($K=2, D=128$) is better than ($K=128, D=8$) in both BLEU and CR, with both DPQ-SX and DPQ-VQ. Secondly, increasing $K$ or $D$ would typically improve the task performance at the expense of lower CRs, which means one can adjust $K$ and $D$ to achieve the best task performance and compression ratio trade-off. Thirdly, we note that decreasing $D$ has a much more traumatic effect on DPQ-VQ than on DPQ-SX in terms of task performance. This is because as the dimension of each sub-space ($d/D$) increases, the nearest neighbour approximation (that DPQ-VQ relies on) becomes less exact.

\subsection{Computational Cost}

DPQ incurs a slightly higher computational cost during training and no extra cost at inference. Figure \ref{fig:train_efficiency} shows the training speed as well as the (GPU) memory required when using DPQ on the medium LSTM model, trained on Tesla-V100 GPUs. For most $K$ and $D$ values, the extra training time is within 10\%, and the extra training memory is zero. For very large $K$ and $D$ values, DPQ-VQ has better computational efficiency than DPQ-SX (as expected). At inference, we do not observe any impact on speed or memory from DPQ.

 \section{Related Work}

Modern neural networks have many parameters and redundancies. The compression of such models has attracted many research efforts~\citep{han2015deep,howard2017mobilenets,chen2018adaptive}. Most of these compression techniques focus on the weights that are shared among many examples, such as convolutional and dense layers~\citep{howard2017mobilenets,chen2018adaptive}. The embedding layers are different in the sense that they are tabular and very sparsely accessed, i.e. the pruning cannot remove rows/symbols in the embedding table, and only a few symbols are accessed in each data sample. This makes the compression challenges different for the embedding layers. Other approaches such as scalar and product quantization~\cite{jegou2010product,joulin2017bag}, as well as compression selection criterion~\cite{may2019downstream} are also explored.

Existing work on compressing embedding layers includes~\citep{shu2017compressing,chen2018learning}, which also leverage discrete codes. As mentioned in Section~\ref{sec:introduction}, our proposed framework is more general and flexible, allowing for two approximation techniques to be used in a single-stage training process. The product keys and values in our model make it more efficient in both training and inference. Empirically, DPQ achieves significantly better compression ratios and at the same time does not need an extra distillation process.

Our work differs from traditional quantization techniques~\citep{jegou2010product} in that they can be trained in an end-to-end fashion. The idea of utilizing multiple orthogonal subspaces/groups for quantization is used in product quantization~\citep{jegou2010product,norouzi2013cartesian} and multi-head attention~\citep{vaswani2017attention}.
 
The two approximation techniques presented for DPQ in this work also share similarities with Gumbel-softmax~\citep{jang2016categorical} and VQ-VAE~\citep{van2017neural}. However, we do not find using stochastic noises (as in Gumbel-softmax) useful since we aim to get deterministic codes. It is also worth pointing out that these techniques~\citep{jang2016categorical,van2017neural} by themselves cannot be directly applied to compression. \section{Conclusion}
In this work, we propose a simple and general differentiable product quantization framework for learning compact embedding layers. We provide two instantiations of our framework, which can readily serve as a drop-in replacement for existing embedding layers. We empirically demonstrate the effectiveness of the proposed method in a wide variety of language tasks. Beyond embedding compression, the proposed DPQ layer can also be used for end-to-end learning of general discrete codes in neural networks.

\section*{Acknowledgements}
We would like to thank Koyoshi Shindo, Mohammad Norouzi, Wang-cheng Kang, Derek Cheng, Dong Lin, Xinyang Yi, Alex Beutel for helpful discussions.

\bibliography{content/ref}
\bibliographystyle{icml2020}

\clearpage
\onecolumn
\appendix

\section{Proof of Proposition \ref{theorem}}
\label{app:proof}

\begin{proof}
We first re-parameterize both the codebook $\mathbf{C}$ and the Value matrix $\mathbf{V}$ as follows.

The original codebook is $\mathbf{C}\in \{1, \cdots, K\}^{n \times D}$, and we turn each code bit, which is an integer in $\{1, \cdots, K\}$, into a small one-hot vector of length-$K$. This results in the new binary codebook $\mathbf{B}\in \{0, 1\}^{n\times KD}$. Per our constraint in proposition \ref{theorem}, $\mathbf{B}$ is a full rank matrix.

The original Value matrix is $\mathbf{V} \in \mathbb{R}^{K \times d}$, and we turn it into a block-diagonal matrix $\mathbf{U}\in \mathbb{R}^{KD \times d}$ where the $j$-th block-diagonal is set to $\mathbf{V}^{(j)}\in \mathbb{R}^{K\times (d/D)}$. Given that each block diagonal, i.e. $\mathbf{V}^{(j)}$, is full rank, the resulting block diagonal matrix $\mathbf{U}$ is also full rank.

With the above re-parameterization, we can write the output embedding matrix $\mathbf{H} = \mathbf{B}\mathbf{U}$. Given both $\mathbf{B}$ and $\mathbf{U}$ are full rank and $KD \ge d$, the resulting embedding matrix $\mathbf{H}$ is also full rank.
\end{proof}

\section{Details of Model Training}
\label{app:datasets_and_models}

We follow the training settings of the base models used, and most of the time, just tune the DPQ hyper-parmeters such as $K$, $D$ and/or subspace-sharing. 

\paragraph{Transformer on WMT'19 En-De.}
For training the Transformer Model on WMT'19 En-De dataset, the training set contains approximately 27M parallel sentences. We generated a vocabulary of 32k sub-words from the training data using the SentencePiece tokenizer~\citep{kudo2018sentencepiece}. The architecture is the Transformer Base configuration described in~\cite{vaswani2017attention} with a context window size of 256 tokens. All models were trained with a batch size of 2048 sentences for 250k steps, and with the SM3 optimizer~\citep{anil-arxiv-2019} with momentum 0.9 and a quadratic learning rate warm-up schedule with 10k warm-up steps. We searched the learning rate in $\{0.1, 0.3\}$.

\paragraph{BERT pre-training.}
As our baseline, we pre-train BERT-base~\citep{devlin2018bert} on 512-token sequences for 1M iterations with batch size 1024. We used the same optimizer (Adam) and learning rate schedule as described in \cite{devlin2018bert}. For the DPQ experiments, we used DPQ-SX with no subspace-sharing, $D=128$ and $K=32$, and exactly the same configurations and hyperparameters as in our baseline.

\section{Code Study}

\subsection{Code Distribution}
\label{app:code_distribution}

\begin{figure}[h]
    \centering
    \begin{subfigure}{0.3\textwidth}
        \centering
        \includegraphics[trim=85 40 135 30,clip,scale=0.3]{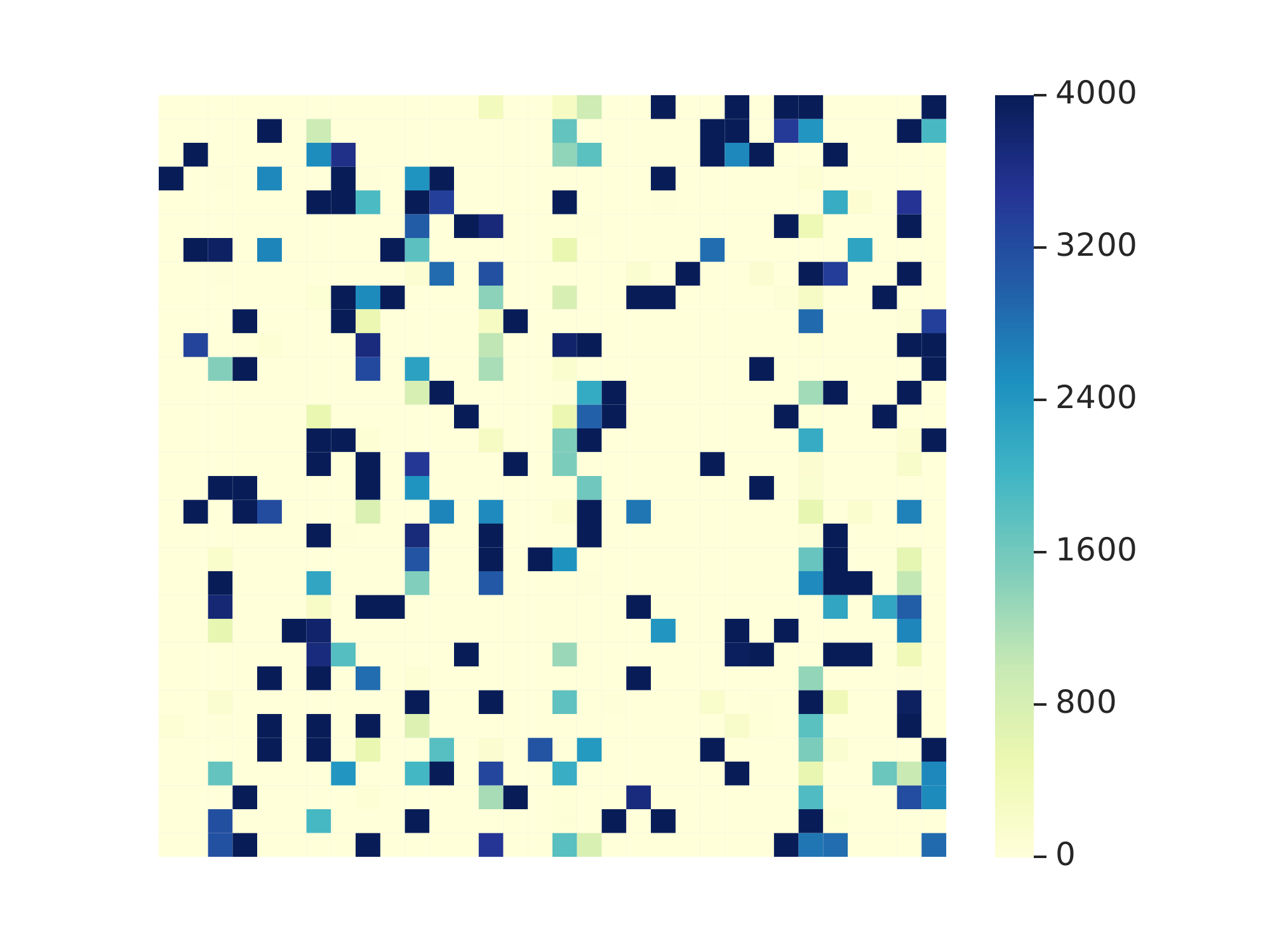}
\end{subfigure}
    \begin{subfigure}{0.3\textwidth}
        \centering
        \includegraphics[trim=85 40 60 30,clip,scale=0.3]{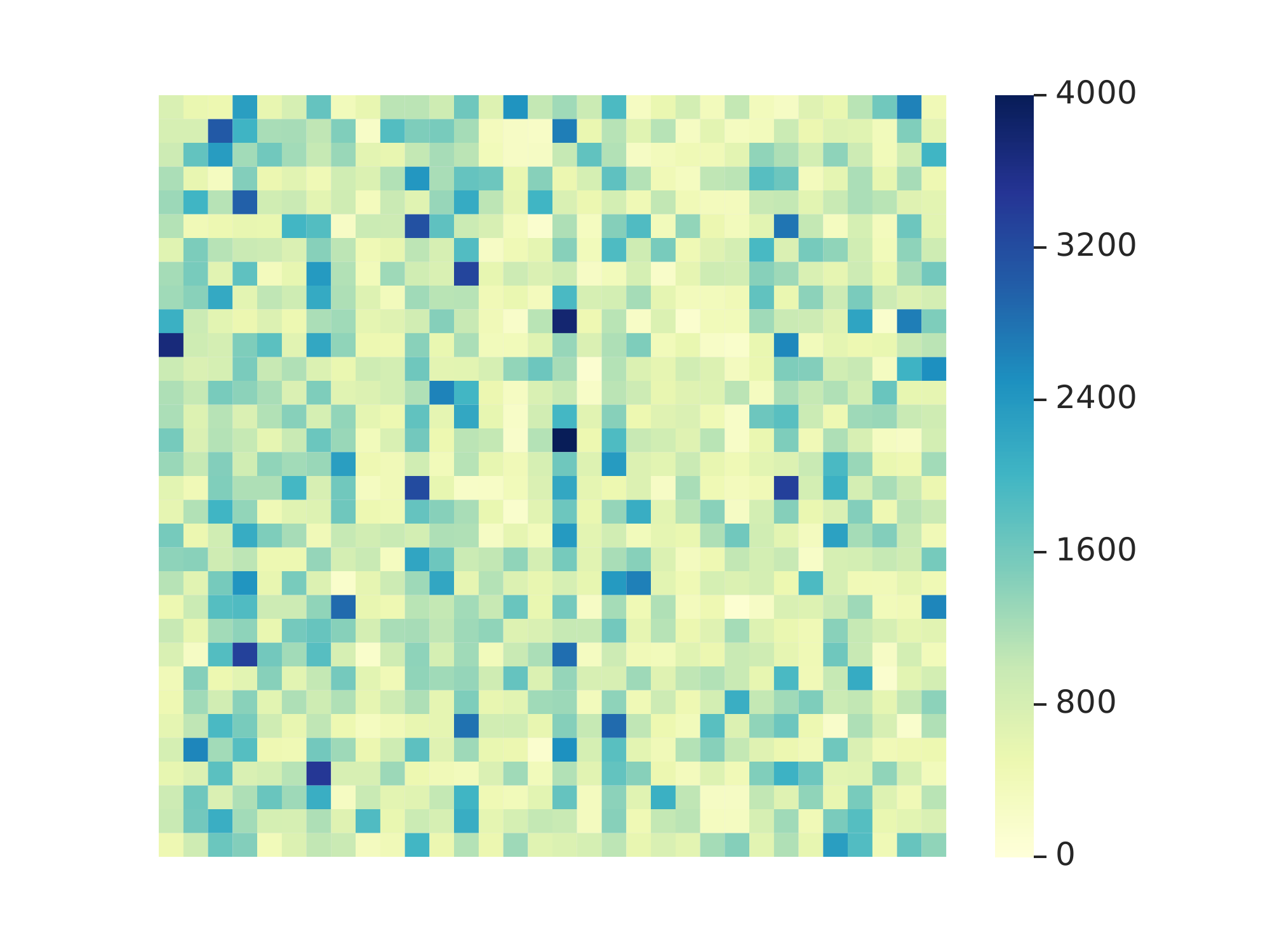}
\end{subfigure}
    \caption{Code heat-maps. Left: DPQ-SX. Right: DPQ-VQ. $x$-axis: K codes per group. $y$-axis: D groups. $K=D=32$.}
    \label{fig:code_heatmap}
\end{figure}

DPQ discretizes the embedding space into the KD codebook in $\{1,...,K\}^{n\times D}$. We examine the code distribution by computing the number of times each discrete code in each of the $D$ groups is used in the entire codebook:
$$
\text{Count}_k^{(j)} = \sum_{i=1}^n (\mathbf{C}_i^{(j)} == k)
$$
Figure \ref{fig:code_heatmap} shows the code distribution heat-maps for the Transformer model on WMT'19 En-De, with $K=32$ and $D=32$ and no subspace-sharing. We find that 1) DPQ-VQ has a more evenly distributed code utilization, 2) DPQ-SX has a more concentrated and sparse code distribution: in each group, only a few discrete codes are used, and some codes are not used in the codebook.

\subsection{Rate of Code Changes}
\label{app:code_change}

\begin{figure}[h]\centering
    \includegraphics[trim=25 10 25 30,clip,width=0.4
    \textwidth]{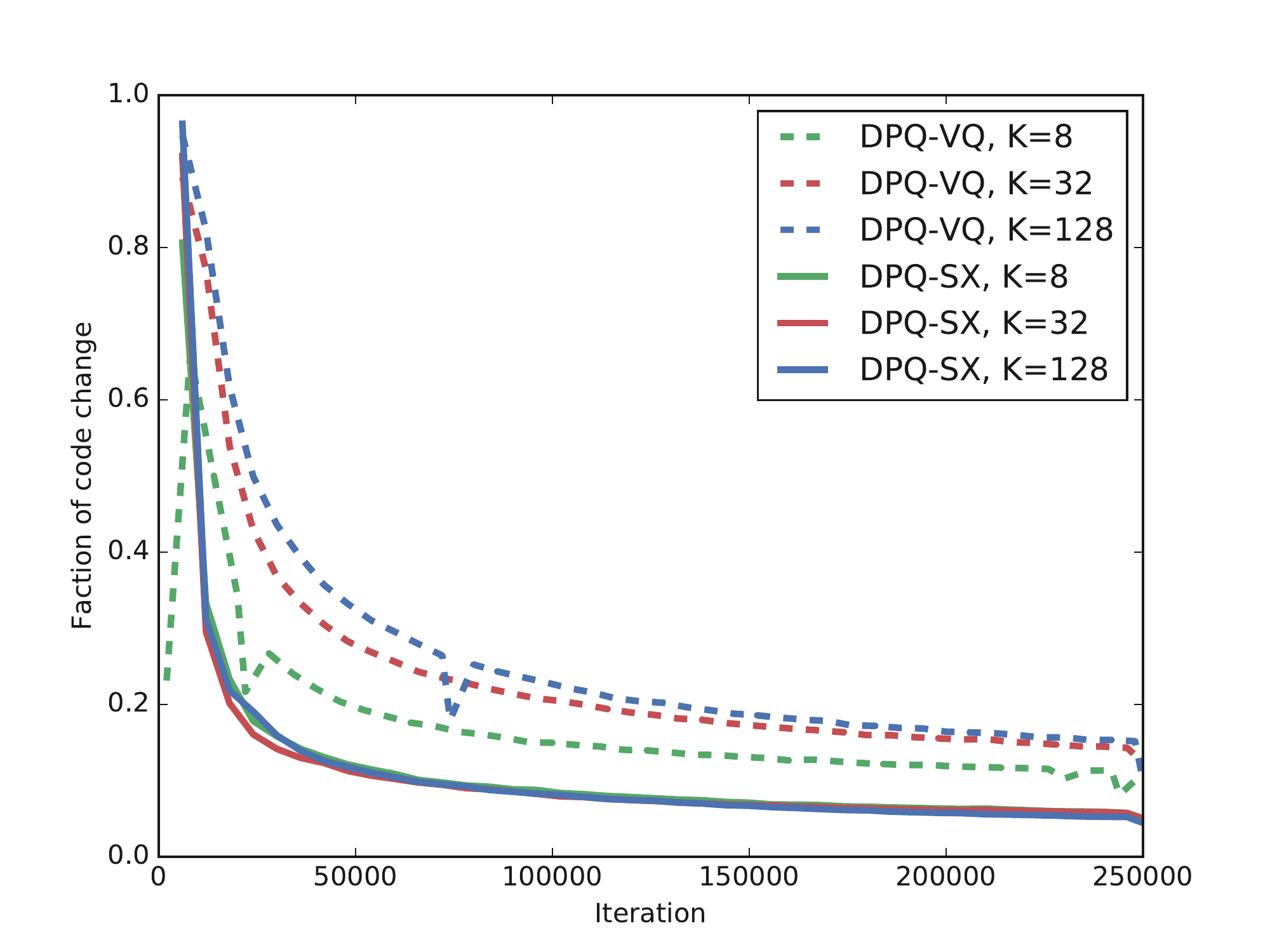}
    \caption{Percentage of code bits in codebook which changed from the previous checkpoint. Transformer on WMT'19 En-De. $D=128$ for all runs. Checkpoints are saved every 600 iterations.}
    \label{fig:code_change}
\end{figure}

We investigate how the codebook changes during training by computing the percentage of code bits in the KD codebook $\mathbf{C}$ changed since the last saved checkpoint. An example is plotted in Figure \ref{fig:code_change} for the Transformer on WMT'19 En-De task, with $D=128$ and various $K$ values. Checkpoints were saved every 600 iterations. Interestingly, for DPQ-SX, code convergence remains about the same for different $K$ values; while for DPQ-VQ, the codes takes longer to stabilize for larger $K$ values.

\subsection{Nearest Neighbours of Reconstructed Embeddings}
\label{app:nearest}

\begin{table*}[ht]
\caption{Nearest neighbours of `\_evolve' in the embedding space.}
\label{tab:evolve}
\begin{center}
\begin{small}
\begin{tabular}{lrlrlr}
\toprule
\textbf{Baseline (Full)} & \textbf{Dist} & \textbf{DPQ-SX} & \textbf{Dist} & \textbf{DPQ-VQ} & \textbf{Dist} \\
\midrule
\_evolve	&	1.000	&	\_evolve	&	1.000	&	\_evolve	&	1.000	\\
\_evolved	&	0.533	&	\_evolved	&	0.571	&	\_evolved	&	0.506	\\
\_evolving	&	0.493	&	\_evolution	&	0.499	&	\_develop	&	0.417	\\
\_develop	&	0.434	&	\_develop	&	0.435	&	\_evolving	&	0.359	\\
\_evolution	&	0.397	&	\_evolving	&	0.418	&	\_developed	&	0.320	\\
\_developed	&	0.379	&	\_arise	&	0.405	&	\_development	&	0.307	\\
\_developing	&	0.316	&	\_developed	&	0.405	&	\_developing	&	0.299	\\
\_arise	&	0.298	&	\_resulted	&	0.394	&	\_evolution	&	0.282	\\
\_unfold	&	0.294	&	\_originate	&	0.361	&	\_changed	&	0.278	\\
\_emerge	&	0.290	&	\_result	&	0.359	&	\_grew	&	0.273	\\
\bottomrule
\end{tabular}
\end{small}
\end{center}
\end{table*}

\begin{table*}[ht]
\caption{Nearest neighbours of `\_monopoly' in the embedding space.}
\label{tab:monopoly}
\begin{center}
\begin{small}
\begin{tabular}{lrlrlr}
\toprule
\textbf{Baseline} & \textbf{Dist} & \textbf{DPQ-SX} & \textbf{Dist} & \textbf{DPQ-VQ} & \textbf{Dist} \\
\midrule
\_monopoly	&	1.000	&	\_monopoly	&	1.000	&	\_monopoly	&	1.000	\\
\_monopolies	&	0.613	&	\_monopolies	&	0.762	&	\_monopolies	&	0.509	\\
monopol	&	0.552	&	monopol	&	0.714	&	monopol	&	0.483	\\
\_Monopol	&	0.380	&	\_Monopol	&	0.531	&	\_Monopol	&	0.341	\\
\_moratorium	&	0.271	&	\_zugestimmt	&	0.486	&	\_dominant	&	0.258	\\
\_privileged	&	0.269	&	legitim	&	0.420	&	\_moratorium	&	0.239	\\
\_unilateral	&	0.262	&	\_Großunternehmen	&	0.401	&	\_autonomy	&	0.230	\\
\_miracle	&	0.260	&	\_Eigenkapital	&	0.400	&	\_zugelassen	&	0.227	\\
\_privilege	&	0.254	&	\_wirkungsvoll	&	0.399	&	\_imperial	&	0.226	\\
\_dominant	&	0.250	&	\_UCLAF	&	0.388	&	\_capitalist	&	0.223	\\
\bottomrule
\end{tabular}
\end{small}
\end{center}
\end{table*}

\begin{table*}[!ht]
\caption{Nearest neighbours of `\_Toronto' in the embedding space.}
\label{tab:toronto}
\begin{center}
\begin{small}
\begin{tabular}{lrlrlr}
\toprule
\textbf{Baseline} & \textbf{Dist} & \textbf{DPQ-SX} & \textbf{Dist} & \textbf{DPQ-VQ} & \textbf{Dist} \\
\midrule
\_Toronto	&	1.000	&	\_Toronto	&	1.000	&	\_Toronto	&	1.000	\\
\_Vancouver	&	0.390	&	\_Chicago	&	0.475	&	\_Orlando	&	0.307	\\
\_Tokyo	&	0.378	&	\_Orleans	&	0.467	&	\_Detroit	&	0.306	\\
\_Ottawa	&	0.372	&	\_Melbourne	&	0.435	&	\_Canada	&	0.280	\\
\_Philadelphia	&	0.353	&	\_Miami	&	0.434	&	\_London	&	0.280	\\
\_Orlando	&	0.345	&	\_Vancouver	&	0.415	&	\_Glasgow	&	0.276	\\
\_Chicago	&	0.340	&	\_Tokyo	&	0.407	&	\_Montreal	&	0.272	\\
\_Canada	&	0.330	&	\_Ottawa	&	0.405	&	\_Vancouver	&	0.271	\\
\_Seoul	&	0.329	&	\_Azeroth	&	0.403	&	\_Philadelphia	&	0.267	\\
\_Boston	&	0.325	&	\_Antonio	&	0.400	&	\_Hamilton	&	0.264	\\
\bottomrule
\end{tabular}
\end{small}
\end{center}
\end{table*}

Table \ref{tab:evolve}, \ref{tab:monopoly} and \ref{tab:toronto} show examples of nearest neighbours in the reconstructed continuous embedding space, trained in the Transformer model  on the WMT'19 En-De task. Distance between two sub-words is measured by the cosine similarity of their embedding vectors. Baseline is the original full embeddings model. DPQ variants were trained with $K=D=128$ with no subspace-sharing.

Taking the sub-word `\_evolve' as an example, DPQ variants give very similar top 10 nearest neighbours as the original full embedding: both have 7 out of 10 overlapping top neighbours as the baseline model. However, in DPQ-SX the neighbours have closer distances than the baseline, hence a tighter cluster; while in DPQ-VQ the neighbours are further from the original word. We observe similar patterns in the other two examples.

\subsection{Code Visualization}
\label{app:code_visualization}

Table \ref{tab:kd_codes} shows some examples of compressed codes for both DPQ-SX and DPQ-VQ. Semantically related words share common codes in more dimensions than unrelated words.

\begin{table*}[!h]
\caption{Examples of KD codes.}
\label{tab:kd_codes}
\begin{center}
\begin{small}
\begin{tabular}{l p{0.5cm} llllllll p{0.5cm} llllllll}
\toprule
&& \multicolumn{8}{c}{DPQ-SX} && \multicolumn{8}{c}{DPQ-VQ} \\
\midrule
\_Monday& &2 &5 &0 &7 &0 &6 &1 &6& &6 &5 &0 &2 &4 &3 &1 &7 \\
\_Tuesday& &6 &0 &0 &7 &0 &6 &1 &7& &1 &7 &0 &2 &0 &3 &1 &7 \\
\_Wednesday& &6 &5 &0 &3 &0 &6 &1 &6& &6 &2 &3 &2 &0 &2 &1 &7 \\
\_Thursday& &5 &5 &0 &3 &0 &6 &1 &7& &7 &2 &0 &2 &0 &3 &1 &2 \\
\_Friday& &4 &6 &0 &7 &0 &6 &1 &7& &6 &0 &0 &2 &1 &6 &1 &7 \\
\_Saturday& &4 &0 &6 &7 &0 &6 &1 &0& &6 &2 &0 &2 &3 &3 &1 &7 \\
\_Sunday& &2 &0 &0 &3 &0 &6 &1 &6& &7 &2 &0 &2 &6 &3 &1 &7 \\
\midrule
\_Obama& &2 &6 &7 &2 &5 &7 &3 &7& &2 &3 &1 &6 &6 &1 &7 &4 \\
\_Clinton& &2 &4 &7 &2 &3 &5 &6 &7& &5 &3 &5 &6 &6 &0 &7 &4 \\
\_Merkel& &4 &1 &7 &2 &6 &2 &2 &6& &6 &3 &1 &1 &4 &6 &7 &4 \\
\_Sarkozy& &7 &6 &7 &1 &4 &2 &5 &0& &0 &3 &1 &7 &5 &7 &7 &4 \\
\_Berlusconi& &4 &6 &5 &1 &4 &2 &6 &7& &6 &3 &0 &6 &6 &7 &7 &4 \\
\_Putin& &2 &6 &7 &1 &6 &7 &6 &7& &5 &3 &1 &6 &6 &7 &7 &6 \\
\_Trump& &7 &6 &7 &2 &0 &7 &6 &7& &2 &3 &1 &6 &5 &7 &7 &7 \\
\midrule
\_Toronto& &6 &2 &3 &2 &4 &2 &2 &6& &4 &3 &4 &7 &6 &2 &0 &7 \\
\_Vancouver& &2 &1 &3 &2 &6 &2 &5 &6& &7 &3 &6 &6 &6 &2 &3 &1 \\
\_Ottawa& &2 &5 &6 &1 &6 &2 &2 &7& &6 &3 &1 &6 &6 &2 &0 &4 \\
\_Montreal& &4 &0 &0 &2 &6 &2 &1 &7& &4 &3 &1 &1 &6 &2 &0 &1 \\
\_London& &1 &2 &0 &2 &4 &7 &1 &7& &2 &3 &0 &2 &6 &3 &3 &7 \\
\_Paris& &4 &0 &3 &5 &4 &2 &1 &0& &5 &3 &0 &0 &6 &3 &2 &7 \\
\_Munich& &4 &2 &0 &4 &0 &7 &5 &0& &1 &3 &3 &5 &6 &3 &1 &7 \\
\bottomrule
\end{tabular}
\end{small}
\end{center}
\end{table*}

\end{document}